\documentclass[final]{cvpr}

\usepackage{times}
\usepackage{epsfig}
\usepackage{graphicx}
\usepackage{amsmath}
\usepackage{amssymb}

\usepackage[pagebackref=true,breaklinks=true,colorlinks,bookmarks=false]{hyperref}
\usepackage[utf8]{inputenc} 
\usepackage[T1]{fontenc}    
\usepackage{hyperref}       
\usepackage{url}            
\usepackage{booktabs}       
\usepackage{amsfonts}       
\usepackage{nicefrac}       
\usepackage{microtype}      
\usepackage{times}
\usepackage{hyperref}
\usepackage{url}
\usepackage{amsfonts,amssymb}
\usepackage{graphicx}
\usepackage{subfig}
\usepackage{natbib}
\setcitestyle{numbers,square,comma}
\usepackage{bm}
\usepackage{multirow}
\usepackage{colortbl}
\usepackage{hyperref}
\usepackage{enumitem}
\usepackage{wrapfig,,lipsum,booktabs}
\usepackage{placeins}
\graphicspath{{figures/}}
\usepackage[
  separate-uncertainty = true,
  multi-part-units = repeat
]{siunitx}
\usepackage[normalem]{ulem}
\newcommand{\tens}[1]{{\mathcal{#1}}}

\iffalse
\newcommand{\strikethrough}[1]{{\color{red}\sout{#1}}}
\newcommand{\update}[1]{{\color{blue}{#1}}}
\newcommand{\sheng}[1]{{\color{blue}[\textbf{\sc Sheng}: \textit{#1}]}}
\newcommand{\ruoming}[1]{{\color{green}[\textbf{\sc Ruoming}: \textit{#1}]}}
\newcommand{\quoc}[1]{{\color{red}[\textbf{\sc Quoc}: \textit{#1}]}}
\newcommand{\mingxing}[1]{{\color{blue}[\textbf{\sc Mingxing}: \textit{#1}]}}
\newcommand{\norm}[1]{{\color{green}[\textbf{\sc Norm}: \textit{#1}]}}
\newcommand{\liqun}[1]{{\color{red}[\textbf{\sc Liqun}: \textit{#1}]}}
\newcommand{\andrew}[1]{{\color{red}[\textbf{\sc Andrew}: \textit{#1}]}}
\else
\newcommand{\sheng}[1]{}
\newcommand{\ruoming}[1]{}
\newcommand{\quoc}[1]{}
\newcommand{\mingxing}[1]{}
\newcommand{\norm}[1]{}
\newcommand{\liqun}[1]{}
\newcommand{\andrew}[1]{}
\newcommand{\update}[1]{{{#1}}}
\newcommand{\strikethrough}[1]{}
\fi

\DeclareMathOperator*{\argmax}{arg\,max}

\def\cvprPaperID{10624} 
\def\confYear{CVPR 2021}

\begin{document}

\title{Searching for Fast Model Families on Datacenter Accelerators}
\author{%
  Sheng Li, Mingxing Tan, Ruoming Pang, Andrew Li, Liqun Cheng, Quoc Le, Norman P. Jouppi  \\
  Google \\
  \texttt{\{lsheng,tanmingxing,rpang,andrewyli,liquncheng,qvl,jouppi\}@google.com} \\
}

\maketitle
\begin{abstract}
Neural Architecture Search (NAS), together with model scaling, has shown remarkable progress in designing high accuracy and fast convolutional architecture families. However, as neither NAS nor model scaling considers sufficient hardware architecture details, they do not take full advantage of the emerging datacenter (DC) accelerators. In this paper, we search for fast and accurate CNN model families for efficient inference on DC accelerators. We first analyze DC accelerators and find that existing CNNs suffer from insufficient operational intensity, parallelism, and execution efficiency. These insights let us create a DC-accelerator-optimized search space, with space-to-depth, space-to-batch, hybrid fused convolution structures with vanilla and depthwise convolutions, and block-wise activation functions. 
On top of our DC accelerator optimized neural architecture search space, we further propose a latency-aware compound scaling (LACS), the first multi-objective compound scaling method optimizing both accuracy and latency. Our LACS discovers that network depth should grow much faster than image size and network width, which is quite different from previous  compound scaling.
With the new search space and LACS, our search and scaling on datacenter accelerators results in a new model series named EfficientNet-X. EfficientNet-X is up to more than 2X faster than EfficientNet (a model series with state-of-the-art trade-off on FLOPs and accuracy) on TPUv3 and GPUv100, with comparable accuracy. EfficientNet-X is also up to 7X faster than recent RegNet and ResNeSt on TPUv3 and GPUv100. 

\end{abstract}

\vspace{-0.2in}
\section{Introduction}
\label{intro}

As Moore's Law is slowing down, more specialized datacenter (DC) accelerators such as GPUs~\cite{nv:a100, nv:v100} and TPUs~\cite{tpuv1:isca:2017,TPUv3,dean2019deep} have been developed to keep up with the increasing demand of machine learning (ML) models. With the increasing complexity of ML model architectures and accelerator architectures, there is a fast-widening gap between achieved performance and available performance. 

Neural Architecture Search (NAS)~\cite{nas_cifar17,searchreinforce16,nas_imagenet18, nastransfer18}, a new paradigm of assembling models automatically, has the potential to bridge the gap. 
Modern NAS usually aims at designing a family of models for different accuracy-speed trade-offs for different use cases.  Because of the high cost associated with searching for the entire family of models, model scaling is commonly used to achieve this goal by scaling~\cite{resnet16,efficientnet19} up from a base model to form a model family. However, on specialized DC accelerators the fast-widening gap remains even with NAS and model scaling, 
because they do not have sufficient visibility into hardware architecture details and thus cannot design optimal model families for them.   

\begin{figure}[t!]
    \centering
    \includegraphics[height=43mm, width=0.48\textwidth]{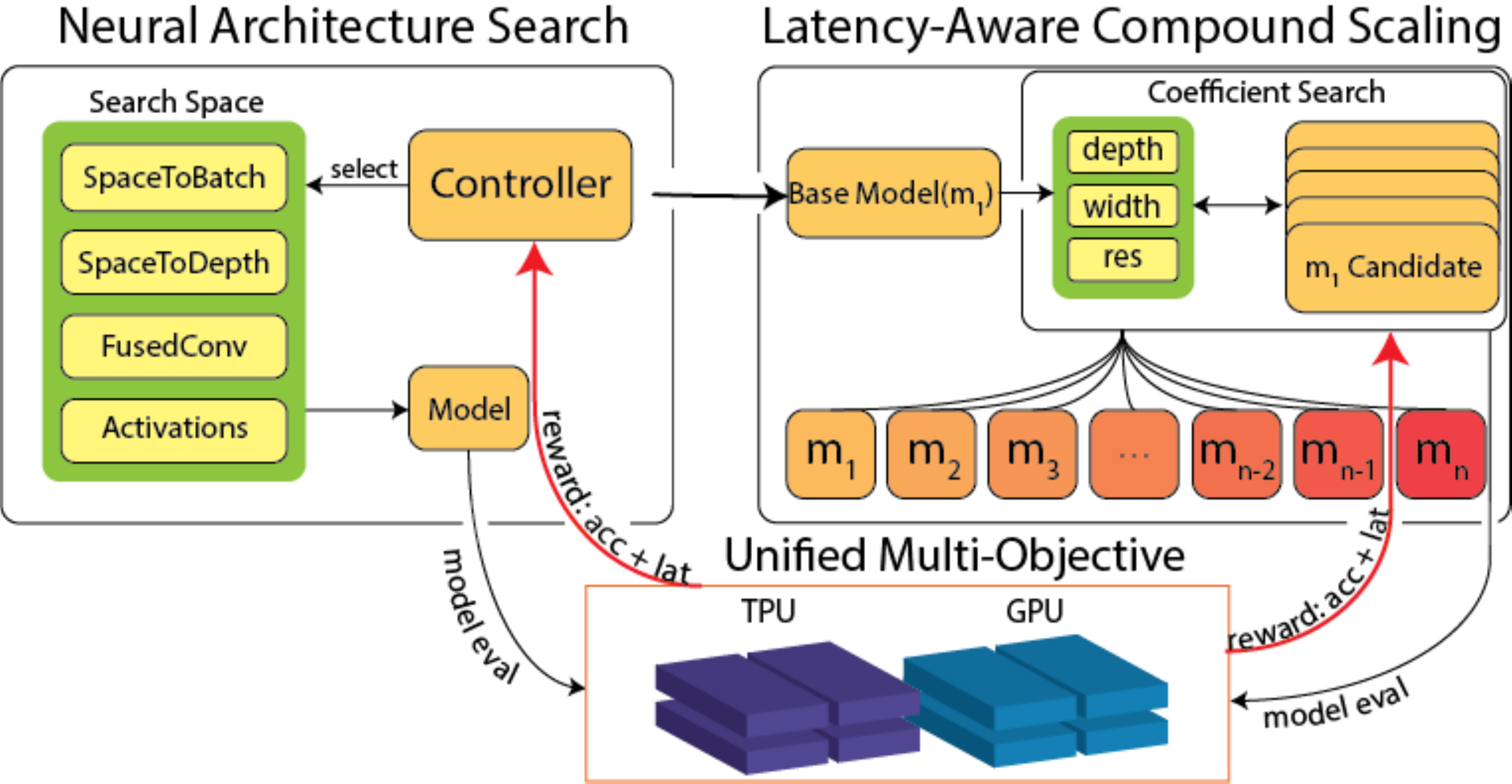} 
    \caption{Unified accelerator-optimized NAS and Latency-aware Compound Scaling (LACS) to search model families optimized for TPUs and GPUs. The same multi-objective with both latency and accuracy is used for both NAS and model scaling. For a given accelerator, a base model (m1) is obtained via NAS with a new search space tailored to DC accelerators. The new latency-aware compound scaling (LACS) searches for scaling coefficients on m1 to form the model family. Both processes are executed separately on TPU and GPU, resulting in two families of final models.}
    \label{fig:nas_paradigm}
    \vspace{-0.1in}
\end{figure}

In this paper, we aim at bridging this gap and designing model families with high accuracy and inference speed, by taking into consideration hardware architecture details of TPUs and GPUs for both NAS and model scaling. We first analyze DC accelerators to find performance bottlenecks. Our analysis reveals the root cause of the recent observed FLOPs-latency nonpropotionality~\cite{ridnik2020tresnet}. We discover that SOTA CNNs suffer from low operational intensity and parallelism, which causes low computation rate (\ie, FLOPs/sec or Ops/sec\footnote{When operations are done in different data types such as bfloat16~\cite{dean2019deep}, float16~\cite{nv:v100}, and tf32~\cite{nv:a100}, the computation rate is usually denoted as OPS, \ie, OPs/Second. Hereafter in this paper, we use FLOPs/sec and Ops/sec interchangeably unless noted otherwise.}) and sub-optimal inference latency and throughput on TPU/GPU accelerators.  With these insights, we augment state-of-the-art (SOTA) NAS with \emph{DC accelerator optimized search space} to improve CNN model operational intensity and efficiency. 
Concretely, we create a new search space with accelerator-friendly operations including space-to-depth, space-to-batch, fused convolution structures, and block-wise searchable activation as shown in Figure~\ref{fig:nas_paradigm}. We propose \emph{latency-aware compound scaling (LACS)} that uses a multi-objective of both accuracy and inference speed to search for scaling factors to generate a model family. LACS is the \emph{first} compound scaling method with a multi-objective including both latency and accuracy. 

With the improved NAS and LACS, we search for high accuracy CNNs for efficient inference on TPUv3~\cite{TPUv3,dean2019deep} and GPUv100~\cite{nv:v100}. Our search results in a new model family named EfficientNet-X (with differences on TPU and GPU) that achieve a better accuracy and latency trade-offs than the state-of-the-art. EfficientNet-X models are up to more than 2X faster on TPUv3 and GPUv100 respectively than EfficientNet~\cite{efficientnet19} with comparable accuracy. Moreover, EfficientNet-X models achieve 30\% more speedup compared to EfficientNet when moving from TPUv2 to TPUv3, demonstrating the generality of our search method across different accelerator generations. EfficientNet-X is also faster than other SOTA models, with on average (geo-mean) 82\% and 48\% faster than RegNet and ResNeSt respectively on GPUv100 and 7X and 48\% faster than RegNet and ResNeSt respectively on TPUv3.

In summary, this paper makes the following contributions:

\begin{enumerate}[noitemsep,topsep=0pt,leftmargin=*] 
    \item We conduct quantitative analysis to reveal the root cause of FLOPs-latency nonproportionality on DC accelerators. Although recent work~\cite{ridnik2020tresnet} has observed the similar behavior, our roofline model and analysis is the \emph{first} to show the fundamental reasons for latency to be much less correlated to FLOPs on GPUs and TPUs than on CPUs. Moreover, our analysis also discovers the performance bottlenecks of CNNs and inspires enhancements for both NAS and compound model scaling. 
    \item We design a DC-accelerator-optimized search space, with space-to-batch, space-to-depth, fused convolution structures, and block-wise activation functions, to compose CNNs with higher operational intensity and efficiency for better accuracy and speed trade-offs.

    \item We propose latency-aware compound scaling (LACS), the \emph{first} compound scaling method with accuracy and latency as the multi-objective. Aftering taking into latency into account, our LACS discovers network depth should grow much faster than image size and network width, which is quite different from previous single-objective compound scaling~\cite{efficientnet19}.
    
    \item Our unified NAS and LACS produce EfficientNet-X, with up to 2X speedup over the EfficientNet and up to 7X speedup over RegNet/ResNeSt on TPUs and GPUs.  
\end{enumerate}

The remainder of this paper is organized as follows. Section~\ref{sec:rethink_speed} provides an analysis on implications of DC accelerator architectures on model performance. Section~\ref{sec:search_space_nas} describes our NAS search space optimized for DC accelerators. Section~\ref{sec:proposal} proposes LACS and integrates it with NAS for an end-to-end search and scaling method for designing model families on DC accelerators. Section~\ref{sec:search_scaling} demonstrates our search and scaling details for composing high accuracy and performant models on TPUs and GPUs, which is followed by Section~\ref{sec:experiments} with experiment results. After describing related work in Section~\ref{sec:related_work}, we conclude in Section~\ref{sec:conclusions}.
\section{Rethink model speed on DC accelerators: Why FLOPs and latency do not correlate}
\label{sec:rethink_speed}
Emerging datacenter accelerators, including TPUs~\citep{tpuv1:isca:2017,dean2019deep} and GPUs, have been using new hardware architectures to keep up with the fast-increasing demand of computing power from ML models. In particular, because matrix-multiplication is the core operation in neural networks, the most special feature of these accelerators is the matrix-multiply-and-accumulate units, called tensor cores~\citep{nv:v100} in GPUs and matrix multiply units~\citep{tpuv1:isca:2017,dean2019deep} in TPUs. These new hardware architectures have changed the way ML models execute on the accelerators. For example, recent work~\cite{ridnik2020tresnet} has observed that FLOPs and latency do not correlate on these accelerators. However, with these empirical observations, there is yet no in-depth analysis to reveal the root cause.  

\begin{figure}
\includegraphics[width=0.5\textwidth]{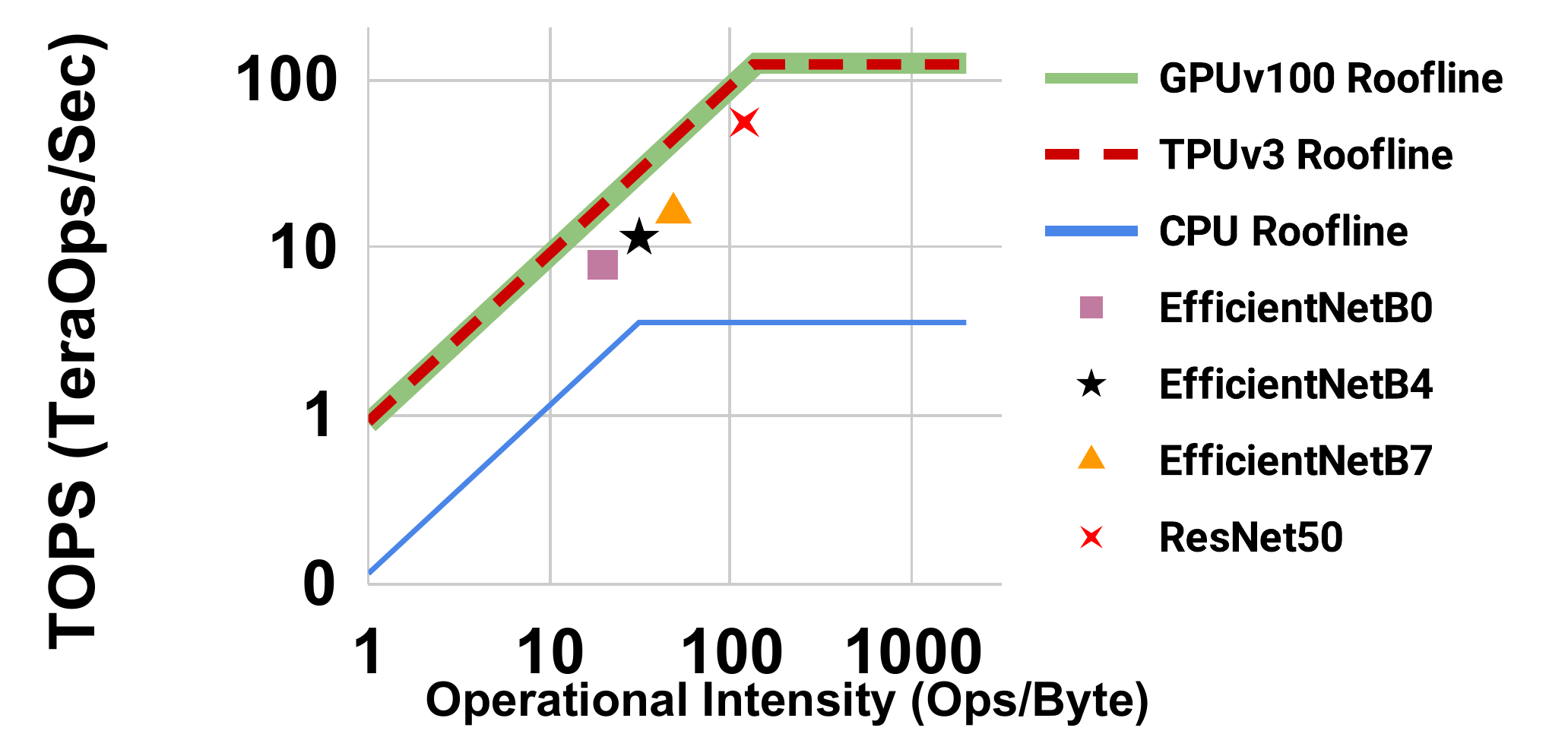}
\caption{\small Rooflines of TPUv3, Volta SMX2 GPU, and Xeon Skylake CPU. TPU and GPU have overlapped rooflines because of their similar peak computation rate and memory bandwidth.  
}
\label{fig:roofline_overall}
\vskip -0.2in
\end{figure}

In this section, we find the root cause of the FLOPs-latency nonpropotionality and provide principles for designing high speed ML models on DC accelerators. 
To rethink the implications of the DC accelerators on model speed including the FLOPs-latency nonpropotionality, we build a generic performance model as shown in Equation~\ref{eqn:optimization}. 
\begin{equation}
\label{eqn:optimization}
\small
\begin{split}
\textbf{Latency} = \frac{W}{C} = \frac{\textbf{W}}{C_{ideal} \times \textbf{E}}, 
\qquad
\textbf{I} = \frac{\textbf{W}}{Q}
\\
C_{ideal} = 
\begin{cases}
  \textbf{I} \times b & \text{if } I < \text{Ridge Point} \\
  C_{max} & \text{else}
\end{cases}
\\
\end{split}
\end{equation}
where $\emph{\textbf{W}}$ (in FLOPs) is the amount of computation required by an ML model, $Q$ (in Bytes) is the memory traffic (bytes of memory transfers) incurred during the execution, and $\emph{\textbf{I}}$ is the operational intensity of the model (in FLOP/Byte). $C$ (in FLOPs/sec) is the computation rate determined by the ideal computation rate ($C_\text{ideal}$) and the execution efficiency $\emph{\textbf{E}}$, where $C_\text{ideal}$ is determined by $I$, accelerator memory bandwidth $b$, and accelerator peak computation rate $C_{max}$. Note that $b$ and $C_{max}$ are accelerator hardware constants. Details of $I$ and $C$ are shown in Figure~\ref{fig:roofline_overall}. The execution efficiency $\textbf{E}$ is defined as the achieved $C$ / $C_\text{ideal}$. The end-to-end inference latency of a model is a nonlinear function of $W$, $I$, and $E$, instead of only $W$--- the FLOPs. This is the \emph{root cause} of FLOPs-latency nonproportionality. 


To dive deeper into the operational intensity and efficiency, we adapt the simple \emph{roofline} analysis (as shown in Figure~\ref{fig:roofline_overall}) that originated from high-performance computing (HPC)\cite{roofline} and has been used in ML~\citep{roofline:GTC:2020,tpuv1:isca:2017}. The roofline model reasonably assumes that applications are either compute-bound or memory-(bandwidth)-bound as they don’t fit in on-chip memories. The Y-axis is computation rate $C$ in FLOPs/sec or Ops/sec, thus the peak computation rate forms the saturation region of the roofline. The X-axis is operational intensity $I$ in FLOPs/memory byte accessed. The memory bytes include weights, activations, and intermediate values. The slope of the linear part can be easily derived to be memory bandwidth (Bytes/Sec). An ML model can achieve peak FLOPs/sec on the accelerators only when its operational intensity is sufficient to push it into the saturation (\ie, compute-bound) region in the roofline.  Otherwise, the ML model is memory-bandwidth-bound. The ridge point is the transition point from the memory-bandwidth-bound performance region to the compute-bound performance region. With the roofline analysis and understanding of datacenter accelerator architectures, we can obtain a few key principles for designing high speed ML models on DC accelerators:
\begin{itemize}[noitemsep,topsep=0pt,leftmargin=*]
\item Compute is significantly cheaper than previous systems because of the new matrix-multiply-and-accumulate units, which results in the $\sim$35X higher TeraOps/sec of GPUv100 and TPUv3 than typical of CPU as shown as the saturation regions in Figure~\ref{fig:roofline_overall}. 
\item ML models need have high operational intensity on TPUs and GPUs to be in the compute-bound region to reach close-to-peak performance. This is because, for TPUs and GPUs, their peak computation rate (TeraOps/s) grows much faster than memory bandwidth (Bytes/s). Thus, TPUs and GPUs have ridge points farther to the right than CPUs. However, as shown in Figure~\ref{fig:roofline_overall} EfficientNets' operational intensity is an order of magnitude lower than that of the TPU/GPU ridge point (and even ResNet), which is too low to tap into the full potential of the DC accelerators despite their significantly reduced FLOPs. Specifically, EfficientNet has$\sim$10X FLOPs reduction compared to other models such as ResNet at comparable accuracy.
\item Parallelism is critical for high speed models. TPU/GPU accelerators are optimized for throughput with the new matrix/tensor units. These matrix/tensor units require large parallelism to achieve high performance. For example, a convolution operation needs to have adequately sized depth, batch, and spatial dimensions to provide enough parallelism to achieve high execution efficiency on matrix units. Additionally, because many vector/element operations such as activation functions run on vector units (\eg, CUDA cores in GPUs and vector units in TPUs) instead of matrix units, sufficient parallelism between matrix and vector units is also important for ML models to achieve high performance on GPUs and TPUs.    
\end{itemize}
\vskip -0.1in

\section{Optimize search space for DC accelerators}
\label{sec:search_space_nas}

Based on the analysis and optimization principles in the previous section, we optimize NAS to improve operational intensity and parallelism to design fast models. NAS has three pillars: the search algorithms governing the search process, the objectives determining the trade-offs of the search results, and the search space as the key link between model architectures and accelerator architectures. Thus, specializing the search space for DC accelerators is crucial to give NAS more visibility to DC accelerator details.  
Our optimized search space includes three key new components:  accelerator-friendly space-to-depth/batch, fused convolution structures, and block-wise activation functions. 

\subsection{Efficient space-to-depth and space-to-batch} 
As pointed out in Section~\ref{sec:rethink_speed}, convolutions need high parallelism in all dimensions (depth, batch, and spatial) to achieve high speed on TPUs and GPUs. However, insufficient parallelism because of the small depth and batch is the usual cause of low utilization and low performance on matrix units. We augment the search space with accelerator-friendly space-to-depth and space-to-batch ops to increase depth and batch dimensions while keeping the total tensor volume the same. 

For space-to-depth ops, instead of using the memory-copy-reshape based ops provided by frameworks such as TensorFlow~\cite{tensorflow} and Pytorch~\cite{pytorch}, we customize an $n \times n$ convolution with stride-n to perform the space-to-depth operation, reshaping an $H\times W \times C $ tensor to an $H/n \times W/n \times C*n^2 $ tensor. This approach has two advantages: 1) convolutions are much preferred by TPUs and GPUs because of their high operational intensity and execution efficiency; 2) in addition to reshaping the input tensor to improve operational intensity and efficiency, the $n \times n$ convolutions can also be trained to contribute to the model's capacity. 
For space-to-batch ops, we have to use the memory-intensive copy-reshape ops provided by common frameworks~\cite{tensorflow, pytorch}.

\subsection{Fused convolution structures} 
As they are the dominant operations in CNNs, it is important to ensure that convolutions in the search space are optimized for accelerator architectures. As the baseline search space already includes a rich set of convolutions with different types, sizes, and shapes, we augment the search space with fused convolution macro structures.  With 4-mode input tensor $\tens{I}$ and output tensor $\tens{O}$ of $N\times C\times H \times W$\footnote{For simplicity, we assume that 1) input depth ($C_\text{in}$) is the same as output depth ($C_{\text{out}}$), 2) input height and weight ($H_\text{in}$ and $W_\text{in}$) are the same as output height and width ($H_\text{out}$ and $W_\text{out}$), and 3) stride-1 square kernels with size $K \times K$. $N$ is the batch size.}, the total computation load $W$ (in FLOPs) and operational intensity $I$ for convolution and depthwise convolution are in Equation~\ref{eqn:conv}. From Equation~\ref{eqn:optimization} and~\ref{eqn:conv}, it is clear that although depthwise convolutions have fewer FLOPs, they also have lower operational intensity to potentially hurt computation rate and thus hurt latency.  
\vskip -0.2in
\begin{equation}
\small
\begin{split}
W_\text{\_Conv2}= N \times H\times W \times C^{2} \times K ^{2}, \\  I_\text{\_Conv2} = \frac{N \times H\times W \times C^{2} \times K^{2}}{2* N \times H\times W \times C + C^{2} \times K ^{2}} ,\\
W_\text{\_DWConv}= N \times H\times W\times C \times (C + K ^{2}), \\I_\text{\_DWConv} = \frac{N \times H\times W\times C \times (C + K ^{2})}{(4* N \times H\times W \times C + C \times K ^{2} + C^{2})}   
\end{split}
\label{eqn:conv}
\end{equation}

This trade-off is more complicated in convolution structures such as mobile inverted bottleneck conv (MBConv)~\cite{mobilenetv218}, an important convolution structure in the baseline search space. MBConv is a macro block that includes a expansion layer of 1x1 convolutions, a depthwise convolution, and a projection layer of 1x1 convolutions, together with activation, batch normalization, and skip-connections. A fused variant of MBConv (fused MBConv) combines the depthwise convolutions with the expansion or projection layer as a vanilla convolution. These trade-offs involving $W$, $I$, and $E$ (as shown in Equation~\ref{eqn:optimization} and~\ref{eqn:conv}) are too complicated for manual optimization but are well-suited for NAS to explore. \update{Concretely, fused MBConv has higher operational intensity (good for speed) but higher  FLOPs  (bad  for  speed)  than  MBConv. Thus, fused MBConv can possibly be either faster or slower than  MBConv,  depending  on  the  shape  and  size  of  weights and activations of the macro op. Moreover, the fused MBConv andMBConv contribute differently to the final model accuracy. } Thus, we added the fused MBConv into the baseline factorized search space~\cite{efficientnet19}. Although recently NAS for mobile devices~\citep{EfficientNet-EdgeTPU19} also uses a similar op, our work is the first to provide the in-depth analysis and employ such ops into the DC accelerator search spaces. Our search indeed finds a combination of fused MBConv and MBConv to get the models with Pareto-optimized latency and accuracy as shown in Table~\ref{tab:base_model_breakdown}.


\subsection{Block-wise searchable activation functions} 
While activation functions have been studied thoroughly for their impact on accuracy~\cite{swish18,arora2018understanding}, their impact on speed is less well understood. With the high computing capacity on TPUs and GPUs, the FLOPs difference among different activation functions is negligible. However, because of the low operational intensity of all activation functions and the shape of rooflines (Figure~\ref{fig:roofline_overall}) of TPU and GPU, all activation functions are memory-bound~\cite{nvdia:performance:activation} on TPUv3 and GPUv100. These memory-bound activation functions can have large negative performance impact to the end-to-end model speed, because they can drag the overall model into the slope region of the rooflines (where ML model performance is far away from the TPU/GPU peak performance as shown in Figure~\ref{fig:roofline_overall}. 

The most important optimization for activation functions is fusing~\cite{xla, nvdia:performance:activation} an activation function with its associated convolutions to avoid accessing memory just for computing the activation function. Because activation functions (being element-wise operations) usually run on vector units, their execution can be in parallel with the execution of convolutions when convolutions run on matrix unit. In theory, the fused activation functions can be completely hidden by the execution of convolutions. But, in practice, the software stack plays an crucial role for such optimizations, which manifests as important model accuracy and speed trade-offs. 

Therefore, we enhance the baseline factorized search space~\cite{efficientnet19,mnas18} with searchable activation functions, including ReLU and swish. To make the search space manageable, we make the activation searchable at the block level in the factorized search space, \ie, different blocks can have different activation functions but all layers within the same block use the same activation function.    

\section{Latency-aware compound scaling (LACS)}
\label{sec:proposal}

The optimized search space in previous section helps our goal to compose CNN model families with optimal accuracy and inference latency on different DC accelerators as shown in~Figure~\ref{fig:nas_paradigm}. Particularly, our goal can be defined generally with Equation~\ref{eqn:family_search}.
\vspace{-0.1in}
\begin{align}
\max_{\mathcal{S}_{h_j}, m_{i, h_j}} \mathcal{O}\Big(\text{Accuracy}(m_{i, h_j}), \text{Latency}_{h_j}(m_{i, h_j})\Big)
\label{eqn:family_search}
\end{align}
Given a set of $k$ DC hardware accelerators $h_1, h_2, \ldots h_k \in \mathcal{H}$ of accelerators, we aim at searching for a family of models denoted as $m_{1, h_j}, m_{2, h_j}, \ldots m_{n, h_j} \in \mathcal{M}_{h_j}$. Models in $\mathcal{M}_{h_j}$ specialize in different DC architectures in $\mathcal{H}$ and increase in accuracy at the cost of latency to serve different use cases. The search process is governed by the accuracy and latency \textit{multi-objective} $\mathcal{O}$ evaluating all models in the family of $\mathcal{M}_{h_j}$ on accelerator $h_j$. The model family $\mathcal{M}_{h_j}$ is composed with a model search space of $\mathcal{S}_{h_j}$ tailored for a given accelerator $h_j$. In this work, the DC hardware accelerator set $\mathcal{H}$ focuses on TPUs and GPUs.

\label{subsec:lacs}
Even with state-of-the-art NAS and our enhanced search space as described in Section~\ref{sec:search_space_nas}, it is too costly to search an entire family of models. For example, directly searching for the entire EfficientNet family (B0$\sim$B7) is $\sim$100X more expensive than searching for EfficientNet-B0, the base model in the EfficentNet family~\cite{efficientnet19}. Therefore, model scaling is commonly used together with NAS. Model scaling has changed from simple scaling~\cite{resnet16, resnest2020, mobilenetv117} to more sophisticated compound scaling~\cite{efficientnet19}. Compound scaling~\cite{efficientnet19} is essentially a search algorithm as it searches for the best scaling factors for depth, width, and resolution, under a given objective and constraint. However, although the SOTA compound scaling has demonstrated better results than simple scaling, by systematically scaling depth, width, and resolution of CNNs, there is still a major hurdle preventing it from harvesting the full potential of hardware and working optimally with NAS. Concretely, by using accuracy as the sole objective\footnote{Although compound model scaling also uses FLOPs as the constraints of the scaling factors, the model accuracy is the only objective when searching for the compound scaling factors.} during searching for scaling factors, the existing SOTA compound scaling method cannot consider the performance/speed impact for the resulted model families. 

As we seek to design end-to-end model family search as described in Equation~\ref{eqn:family_search} and Figure~\ref{fig:nas_paradigm}, we propose \emph{latency-aware compound scaling (LACS)}. Unlike existing compound scaling that uses accuracy as the sole objective, LACS employs accuracy and latency as the multi-objective when searching for scaling factors of depth, width, and resolution of CNNs for better latency and accuracy trade-offs. Searching for scaling factors with LACS amounts to approximating the solution to the following optimization problem for each accelerator $h_j$:
\vspace{-0.1in}

\begin{equation}
\begin{aligned}
    d_{h_j}, w_{h_j}, r_{h_j} \\ = \argmax_{d, w, r} &\ \mathcal{O}\Big(\text{Accuracy}(m_{i, h_j}), \text{Latency}_{h_j}(m_{i, h_j})\Big)\\
      \text{w.r.t.}  ~~~ & \text{Latency}(G(m_{i, h_j}, d, w, r)) = T_{m_{i+1, h_j}} &
\end{aligned}
\end{equation}
where $d, w, r$ are scaling coefficients for model's depth, width, and input resolution respectively while preserving basic network architecture. $T_{m_{i+1, h_j}}$ is the target latency for the ($i+1$)th model of the family on $h_j$. $d, w, r$ are determined by a compound coefficient $\phi$ to scale the network uniformly: 
\vspace{-0.1in}
\begin{equation} 
\begin{aligned}
d  = \alpha ^ \phi, w = \beta ^ \phi, r   =  \gamma ^ \phi;  &&\text{s.t.    } \alpha \ge 1, \beta \ge 1, \gamma \ge 1 
\end{aligned}
\label{eq:optobj} 
\end{equation}
$\phi$ controls how many more resources are available for model scaling. In the original compound scaling that uses accuracy as the sole objective, $\phi$ means the extra FLOPs for model scaling. Whereas, in our latency-aware compound scaling, $\phi$ means the latency budget for model scaling, with $\alpha, \beta$ and $\gamma$ controlling how the latency budget is allocated to scale depth, width, and resolution, respectively. $\alpha, \beta$ and $\gamma$ can be determined by a grid search. Base original compound scaling has additional constraints from FLOPs, LACS's search space is larger because the use of accuracy and latency as the multi-objective and the FLOPs-latency nonproportionality. LACS is the first multi-objective compound scaling, which enables streamlined integration with multi-objective NAS with the same unified multi-objective reward including both model accuracy and latency as shown in Figure~\ref{fig:nas_paradigm}.

\section{Search and scaling optimized model families on DC accelerator}
\label{sec:search_scaling}
This section describes our process of searching and scaling to design model families on TPUs and GPUs with the unified NAS and LACS. We first use NAS with the new search space tailored for DC accelerators to search for the base model. We then use LACS to find scaling factors to compose model families on TPUs and GPUs. 

\subsection{NAS for base models}
\label{subsec:nas_search}
We use a NAS infrastructure similar to~\citep{mnas18,efficientnet19}, where we employ the same RNN-based controller. We build an infrastructure to retrieve TPU and GPU hardware latency directly during search and run NAS on TPUv3\cite{TPUv3,dean2019deep} and GPUv100~\cite{nv:v100}. \update{We  used  data parallelism  for  distributed training/searching on both TPUs and GPUs. Since the largest model candidate can fit in a single TPU/GPU device, the data parallelism is sufficient for distributed training on a pool of TPUs/GPUs.} 

\update{As an ablation study, we first use the original search space from EfficientNet~\citep{efficientnet19} and inference latency of TPUv3 and GPUv100 instead of total computation load (FLOPs) as the performance signal. However,  our  search  found  no  model  better  than EfficientNet-B0 with ReLU, because the original Efficient-Net search space did not have the TPU/GPU-optimized operations such as space-to-depth/batch, fused MBConv, and searchable  activation  functions.   Thus,  in  the  original  EfficientNet  search  space  without our  TPU/GPU-optimized  operations,  the FLOPs-optimized  models and latency-optimized models converged to the same model architecture as EfficientNet-B0 with ReLU\footnote{Note  that  when  searching  on  the  original  EfficientNet search space, we always used ReLU because the original EfficientNet search space did not support searching for activation functions. In the original EfficientNet~\citep{efficientnet19},  EfficientNet-B0 was  searched with  ReLU  and  manually  set  to  use  Swish  for  all  layers after the search was done}. This observation further necessitates the design of the new search space customized for TPUs and GPUs. } 

We then performance NAS on our proposed new search space as described in Section~\ref{sec:search_space_nas}. We use the same multi-objective reward mechanism as in~\cite{mnas18}. The multi-objective reward combines accuracy and latency as $ACCURACY(m) \times \left[ \frac{LATENCY(m)}{Target}\right]^w$ to approximate the Pareto-optimal results on both accuracy and latency.  The factor $w$ decides the weight of latency in the reward. We re-calibrate the factor $w$ to make the reward design suitable for TPUv3 and GPUv100. Particularly, we use a larger weight factor $w = -0.09$ because model accuracy is less sensitive to latency variations on TPU and GPU platforms than on mobile platforms (original $w = -0.07$ in~\cite{mnas18}). \update{We choose the multiplicative objective function form, the same form as used in the baseline EfficientNet, to ensure fair comparisons. Different objective function forms, such as additive forms\citep{bender2020weight}, can potentially produce even better results, and we will investigate in future work.}

Our search produces EfficientNet-X-B0, a fast network on TPUs and GPUs, as shown in Table~\ref{tab:b0-arch}. The model architecture is mostly the same on both TPUv3 and GPUv100, except the different activation function selections. The EfficientNet-X-B0 demonstrates the impact of the new accelerator-optimized search space, compared to the baseline EfficientNet-B0~\cite{efficientnet19}. 
\emph{Firstly}, a space-to-depth op using convolution-2x2 with stride-2 is inserted before the second stage
, which can improve the channel depth of subsequent layers to improve speed. \emph{Secondly}, EfficientNet-X-B0 uses hybrid MBConv, with fused-MBConv in  stage 4 and 5 and non-fused MBConv in the rest of the stages. 
\emph{Thirdly}, as mentioned, EfficientNet-X-B0 employs different activation function strategy on TPUs and GPUs. On TPUs, EfficientNet-X-B0 uses swish in stages with fused-MBConv but ReLU in stages with MBConv.  On GPUs, EfficientNet-X-B0 selects ReLU for all stages. 
\emph{Lastly}, NAS designs EfficientNet-X-B0 with bigger squeeze-and-excite layers than EfficientNet-B0. 

\begin{table}[t]
	\vskip -0.05in  
	\caption{\small EfficientNet-X-B0 architecture. The architecture includes multiple stages, with each row representing a stage. Each stage includes operators, number of repeated layers denoted as \#L, (input/hidden) resolution, output channel size denoted as \#OC, squeeze-and-excite (SE) ratio~\cite{senet18}, and activation functions denoted as AF. Activation functions differ on TPUs from GPUs. 
	}
  \vskip -0.1in
  \centering   
  \resizebox{1.0\columnwidth}{!}{ 
	\begin{tabular}{c|c|c|c|c|c|c}                                                     
	Stage & Operator  & Resolution      & \#OC & \#L & SE & AF(TPU/GPU)\\
	\midrule                                                              
	1  &    Conv3x3          & $ 224 \times 224 $   & 32 & 1 & N/A & swish/ReLU \\
	2  &    Conv2x2 for reshaping$^{\dagger}$          & $ 112 \times 112 $   & 128 & 1 & N/A & ReLU/ReLU\\
	3 &   MBConv1, k3x3   & $56 \times 56 $  &   64 &1  & 1 & ReLU/ReLU\\
	4 &   Fused MBConv6, k3x3    & $56 \times 56$ & 24   & 2 & 0.5 & swish/ReLU\\
	5 &   Fused MBConv6, k5x5   &  $56 \times 56$ &  40   & 2  & 0.25 & swish/ReLU\\
	6 &   MBConv6, k3x3    & $28 \times 28 $     & 80   &  3 & 0.25 & ReLU/ReLU\\
	7 &   MBConv6, k5x5  & $14 \times 14$  &  112  & 3 & 0.25 & ReLU/ReLU\\
	8 &   MBConv6, k5x5  & $14 \times 14$ &   192  & 4  & 0.25 & ReLU/ReLU\\
	9 & MBConv6, k3x3   & $7 \times 7$ &   320  & 1  & 0.25 & ReLU/ReLU\\
	10 & Conv1x1 \& Pooling \& FC   & $7 \times 7$ & 1280 & 1 & N/A & ReLU/ReLU\\       
\end{tabular}                                                                  
  }  \label{tab:b0-arch}   
\vskip -0.1in
\end{table}

All the new model architectures in EfficientNet-X-B0 show the effectiveness of the DC accelerator optimized search space. We use selection of activation functions as an example to shed more light.  The usage of swish and ReLU in EfficientNet-X-B0 is complete the opposite of mobilenetv3~\cite{mobilenetv319}. MobilenetV3 uses swish only in later layers, because the cost of applying nonlinearity decreases in deeper layers of the network. Note that swish has $\sim$4X more FLOPs than ReLU making it too expensive on mobile platforms. 

However, as describe in Section~\ref{sec:search_space_nas}, because of the high computing capacity of TPUs and GPUs, the FLOPs differences between swish and ReLU are negligible. Instead, activation functions are optimized with fusion and run on vector units in parallel with convolutions that usually run on matrix units. However, the software stack on GPUs only fuses ReLU with associated convolutions but not swish, which leads to significant slow down for GPU models with swish. As a result, EfficientNet-X-B0 on GPU chooses ReLU for all layers. In contrast, since TPU has swish fused with convolutions through XLA~\cite{xla}, EfficientNet-X-B0 uses swish in many layers. We are initially surprised to see the mixed use of swish and ReLU. But our profiling results with Cloud TPU Profiler~\cite{tpuprof} reveal that depthwise convolutions on TPU run on vector units\footnote{Coincidentally, recent experiment~\cite{depthwisegpu} discovers the similar behavior on GPU. Depthwise convolutions run in vector units, \ie, CUDA cores, instead of the tensor cores on GPUs.} instead of matrix units. Thus, severe contention on vector units happens between depthwise convolutions and swish, as swish has 4X more ops than ReLU despite its benefits in improving model accuracy. When searching on TPUs with our new search space, NAS automatically pairs ReLU with stages containing depthwise convolutions to avoid competition on vector units. Appendix~\ref{sec:search-ablation} shows more ablation studies on the searched base model.

\begin{table}[t!]
\vskip -0.05in  
    \caption{\small Comparison of LACS scaling factors with existing SOTA compound scaling using accuracy as the sole objective (\ie., EfficiencNet's scaling factors).  $\alpha$, $\beta$, and $\gamma$ are the base term of the exponential scaling for depth, width, and resolution respectively, as shown in Equation~\ref{eqn:optimization}.  
    }
    \vskip -0.1in
    \centering
    \small
    \begin{tabular}{c||c|c|c}
        Scaling Type & $\alpha$ (depth) & $\beta$ (width) & $\gamma$ (resolution) \\
        \midrule [0.1em] 
        Accuracy-only & 1.2 & 1.1 & 1.15 \\
        LACS on GPU & 1.28 & 1.17 & 1.07 \\
        LACS on TPU & 1.25 & 1.17 & 1.09 \\
    \end{tabular}
    \label{tab:scaling-coefficients}
    \vskip -0.1in
\end{table}

\subsection{Scaling to form model families with LACS}
\label{subsec:scaling_lacs}
With the searched base model EfficientNet-X-B0, we use LACS to search for scaling factors to build the model family. As described in Section~\ref{subsec:lacs}, we perform Pareto frontier search to find best $\alpha, \beta$, and $\gamma$. We start with initial grid search for coefficient triplets of $\alpha, \beta$, and $\gamma$ using the same multi-objective (\ie, $ACCURACY(m) \times \left[ \frac{LATENCY(m)}{Target}\right]^w$ ) as used in NAS when searching for the base model. We then iterate more fine-grained search in the neighborhood of the best candidate triplets. We search on TPUv3 and GPUv100 and find different optimal scaling coefficients as shown in Table~\ref{tab:scaling-coefficients}. 

\update{LACS discovers network depth should grow much faster than image resolution and network width with image resolution growing slowest, which is quite different from the previous SOTA compound scaling using accuracy as the single objective. Faster increase on network depth than on image resolutions can slow down the memory inflation due to activation and intermediate tensors, which improves model speed by making a model more compute bound than memory bound. As shown in Section~\ref{sec:rethink_speed}, DC accelerators prefer models to be compute-bound to achieve high performance.}
\strikethrough{LACS prefers the deeper and wider network architectures with smaller resolutions on both TPUs and GPUs. We believe that the wider network is because the software stack on these accelerators prioritizes the parallelization on channel dimension over the spatial dimension. Thus, wider networks with more parallelism across channel dimension is critical for high speed models. With the resulting smaller resolution from the wider network, the increased depth is a better fit to achieve good accuracy. }

We also perform direct search on TPUv3 and GPUv100 with the same latency target as EfficientNet-X-B1 and find the same model architectures as obtained by LACS, which confirms that LACS can find the same model as the direct multi-objective NAS when given the same latency target, but with much fewer accelerator resources. Appendix~\ref{sec:lacs-ablation} shows more ablation studies on LACS.

\vskip -0.1in
\section{Experiments}
\label{sec:experiments}
\vskip -0.05in

\begin{table*}[t!]
\small
\vskip -0.1in
  \caption{                                                                      
        \small EfficientNet-X inference speed and accuracy results on ImageNet on TPUv3 and GPUv100. ConvNets with similar top-1 accuracy are grouped together. $^{\star}$Original reported model accuracies in papers are used in the comparisons. $^{\dagger}$Following common practices, \#FLOPs refer to \#multiply-and-add operations. $^{\ddagger}$E is the execution efficiency measured on TPUv3, w.r.t to roofline instead of peak hardware FLOPs/sec as shown in Equation~\ref{eqn:optimization}. Only in the compute-bound region as shown in Figure~\ref{fig:roofline_overall}, the roofline and hardware peak hardware FLOPs/sec are the same. $^{\S}$The inference latency are measured for inferencing 128 images on TPUv3 and GPUv100, with mini batch size of 128. All the measured speed is verified to be the same or faster than the reported results in original papers with the same batch size to ensure fair and correct measurements. Note that the results are to demonstrate the effectiveness of our unified search and scaling method on different DC accelerators. And direct comparing TPU and GPU results is not meaningful and beyond the scope of this paper, because we focus on evaluating the model architecture themselves on different DC accelerators and run models directly on both GPUs and TPUs without extra offline model optimizations (\eg, TensorRT~\cite{tensorRT} and model tuning~\cite{reddi2020mlperf}).    
       }                                                                            
  \vskip -0.1in
    \centering   
    \small
    \resizebox{1.0\textwidth}{!}{                                                  
        \begin{tabular}{l|c|cc|ccc}           
        \small
        \multirow{2}{*}{\bf Models}	 &  \multirow{2}{*}{ \bf Acc.$^{\star}$}	  & \multicolumn{1}{c}{\bf\#Params} & \multicolumn{1}{c|}{\bf \#FLOPs$^{\dagger}$} & \multicolumn{1}{c}{\bf I} & \multirow{2}{*}{\bf E$^{\ddagger}$} & \multicolumn{1}{c}{\bf Inference Latency$^{\S}$(ms)} \\
        &  & (\bf Million) & (\bf Billion) & (\bf Ops/Byte) &  & (\bf  TPUv3 / GPUv100)  \\
        \midrule [0.1em]                                           
        EfficientNet-X-B0  &  77.3\%  & 7.6  & 0.91 & 63.8 & 57.3\% & \bf 8.71 / 22.5  \\
        EfficientNet-B0~\cite{efficientnet19}  & 77.3\%  & 5.3  &   0.39 & 19.7& 52.4\% & 13.4 / 38.1\\
        ResNet-50~\cite{resnet16} & 76.0\%   & 26 &  4.1 & 122.5 & 57.2\% & 35.1 / 35.6\\
        RegNetY-800MF \cite{regnet2020} & 76.3\%   & 6.3 &  0.8 &12.7 & 30\% &45.1 / 33.9 \\

        \midrule
         EfficientNet-X-B1  &  79.4\%  &  9.6  &  1.58 & 65.5 & 59.2\% & \bf 13.6 / 34.4  \\ 
        EfficientNet-B1  &  79.2\%  &  7.8  & 0.70  & 21.4 & 51.3\% & 22.3 / 60.5\\
        Inception-v3 \cite{inceptionv316} & 78.8\%  & 24 & 5.7  & 94.6 & 34.5\% & 104.8 /55.6 \\
        RegNetY-4.0GF \cite{regnet2020} & 79.4\%  & 26 & 4.0 & 19.4 & 29.2\% &109.5 / 75.1\\
        
        \midrule
         EfficientNet-X-B2  &  80.0\%  & 10.0  &   2.3 & 73.0 & 54.8\% &\bf 15.7 / 45.5 \\  
        EfficientNet-B2  &  80.3\%  &   9.2  & 1.0 & 24.1 & 48.8\% & 29.8 / 77.2 \\ 
        Inception-v4 \cite{inceptionv417} & 80.0\%  & 48 &  13 & 148.5 & 35.3\% & 75.1 / 119.9\\  
        RegNetY-8.0GF \cite{regnet2020} & 79.9\%  & 39.2 & 8.0 & 27.9 & 32.4\% & 190.5 / 122.1\\
        
         \midrule
		EfficientNet-X-B3  &  81.4\%  & 13.3  &  4.3 & 84.0 & 51.2\% & \bf 31.9 / 66.6 \\ 
		EfficientNet-B3  &  81.7\%  & 12  &  1.8 & 26.1 & 51.3\% & 48.1 / 128.8 \\ 
		
		\midrule
		 EfficientNet-X-B4  &  83.0\%  &  21.6  &  10.4 & 101.5 & 47.7\% & \bf  64.9  / 149.2  \\ 
		EfficientNet-B4  &  83.0\%  &  19  &   4.2 & 31.29 &  47.8\% & 102.6 / 310.7 \\
		 NASNet-A \cite{nas_imagenet18} & 82.7\%  & 89 &  24 & 55.2 & 43.8\% & 269.5 / 481.2\\
		 ResNeSt-101 \cite{resnest2020} & 83.0\%  &  48 & 13 & 71.7 & 28.1\% & 92.3 / 149.4\\
		 
		\midrule
		 EfficientNet-X-B5  &  83.7\%  & 33.4  &  24.4 & 126.1 & 47.8\% & \bf 125.9 / 290.2 \\ 
		EfficientNet-B5  &  83.7\%  & 30  &  9.9 & 39.7 & 46.8\% & 192.5 / 640.1 \\
		 ResNeSt-200 \cite{resnest2020} & 83.9\%  & 70 & 36.3 & 68.7 & 69.9\% & 244.3 / 415.6\\

		\midrule
		  EfficientNet-X-B6  &  84.4\%  & 47.7  &  47 & 167.5 & 36.2\% & \bf 237.6 / 467.2 \\ 
		 EfficientNet-B6  &  84.4\%  & 43  &   19 & 43.9 & 45.0\% & 334.2 / 1040.6 \\
		\midrule
		 EfficientNet-X-B7  &  84.7\%  &  73.2  &   91 & 194.3 & 39.4\% & \bf 433.9 / 847.7 \\ 
		EfficientNet-B7  & 84.7\%  &  66  &   37 & 48.3 & 43.4\% & 621.4 / 1471.3 \\
		ResNeSt-269 \cite{resnest2020} & 84.5\%  & 111  & 77 &72.9 & 70.2\%  & 501.9 / 864.9\\
        \end{tabular}  
        }
    \label{tab:imagenet}     
    \vskip -0.2in
\end{table*}

We present the accuracy and performance results on the new EfficientNet-X model family on TPUs and GPUs, to demonstrate the effectiveness of the unified NAS and LACS method. 
Table~\ref{tab:imagenet} shows the speed and accuracy on ImageNet~\cite{imagenet15} of EfficientNet-X models and comparisons with other SOTA CNN models, where a few key observations can be made. \emph{First}, EfficientNet-X models are the fastest among each model group on TPUs and GPUs, with comparable accuracy. Specifically, EfficientNet-X models are up to more than 2X faster than EfficientNet, with geometric mean speedup of 56\% and 83\% on TPUv3 and GPUv100 respectively. EfficientNet-X is on average (geomean) 82\% and 48\% faster than RegNet and ResNeSt respectively on GPUv100 and 7X and 48\% faster than RegNet and ResNeSt respectively on TPUv3. \emph{Second}, all models except for Efficient-X models in Table~\ref{tab:imagenet} are polarized. 
On one extreme, the EfficientNet family has the fewest FLOPs but the lowest operational intensity $I$. On the other extreme, other models such as ResNet and Inception families have the highest operational intensity but most FLOPs. Note that while lower FLOPs improves inference speed, lower operational intensity hurts inference speed.  In contrast, the EfficientNet-X models strike a balance between computation load and computation rate, having both FLOPs and operational intensity in the middle between the two extremes, which makes EfficientNet-X models to be the fastest in each group. 
\vskip -0.03in

\begin{figure}[t]
\includegraphics[width=0.5\textwidth]{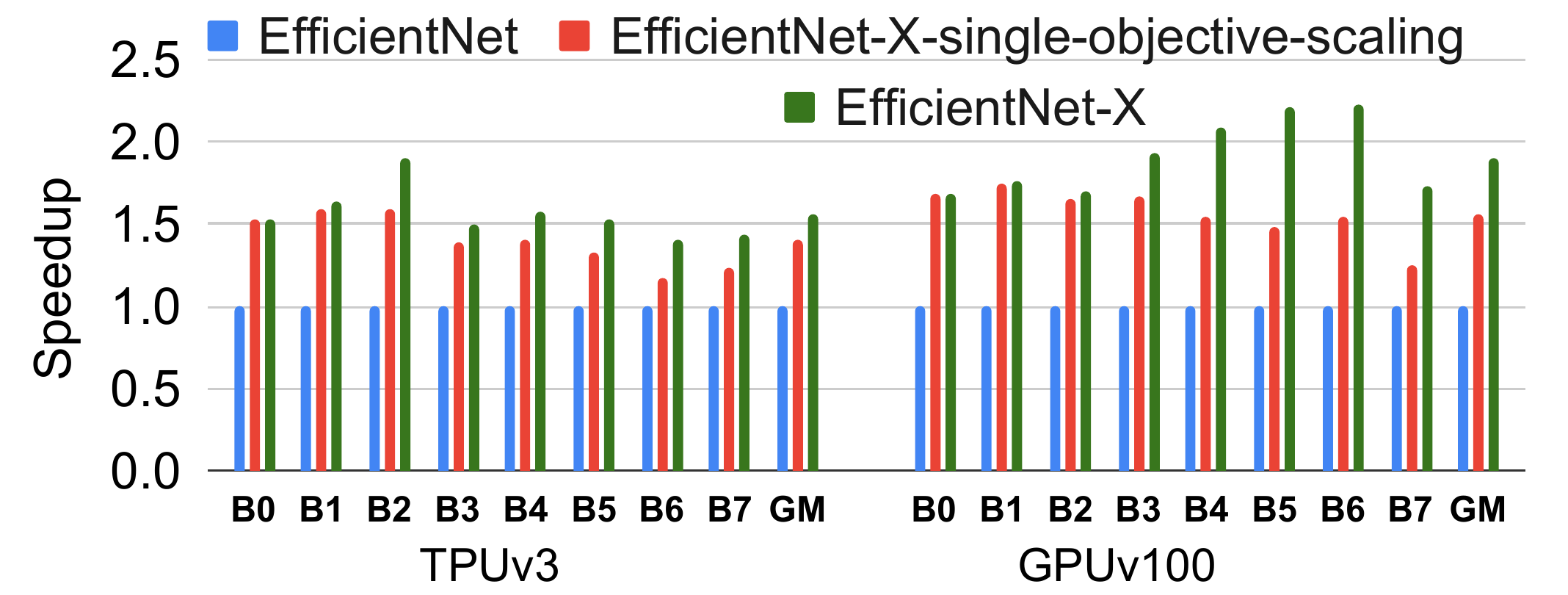}
\caption{\small Speedup of EfficientNet-X and EfficientNet-X-single-objective-scaling over the baseline EfficientNet. EfficientNet-X-single-objective-scaling forms the model family use EfficientNet-X-B0 as the base model but uses original EfficientNet's scaling factors that are obtained by compound scaling with accuracy as the sole objective. GM is geometric mean.
}
\label{fig:speedup_nas_lacs}
\end{figure}

Figure~\ref{fig:speedup_nas_lacs} shows the speedup details due to our new search and scaling method. Overall, EfficientNet-X achieves up to 2X+ speedup on TPUv3 and GPUv100 over EfficientNet, with geometric mean speedup as 56\% and 91\% on TPUs and GPUs respectively. Figure~\ref{fig:speedup_nas_lacs} also shows the ablation study on the speedup breakdown due to NAS with the new search space and LACS. EfficientNet-X-single-objective-scaling composes the model family using EfficientNet-X-B0 as the base model but the EfficientNet's orginal scaling factors that are obtained by single-objective compound scaling with accuracy as the sole objective. Thus, the speedup on EfficientNet-X-B0 over EfficientNet-B0 shows the benefits of the NAS with new search space, and the relative speedup of EfficientNet-X over EfficientNet-X-single-objective-scaling in Figure~\ref{fig:speedup_nas_lacs} indicates the benefits of LACS over previous SOTA compound scaling with accuracy as the only objective. Concretely, NAS with new search space generates $\sim$50\% speedup on TPUv3 and GPUv100, respectively. LACS further increases performance by 14\% and 25\% average (geometric mean) on TPUs and GPUs respectively, atop the speedup due to the new search space. The more detailed ablation studies on search space and LACS can be found in Appendix~\ref{sec:search-ablation} and ~\ref{sec:lacs-ablation} respectively.


\begin{figure}[t]
\includegraphics[width=0.48\textwidth]{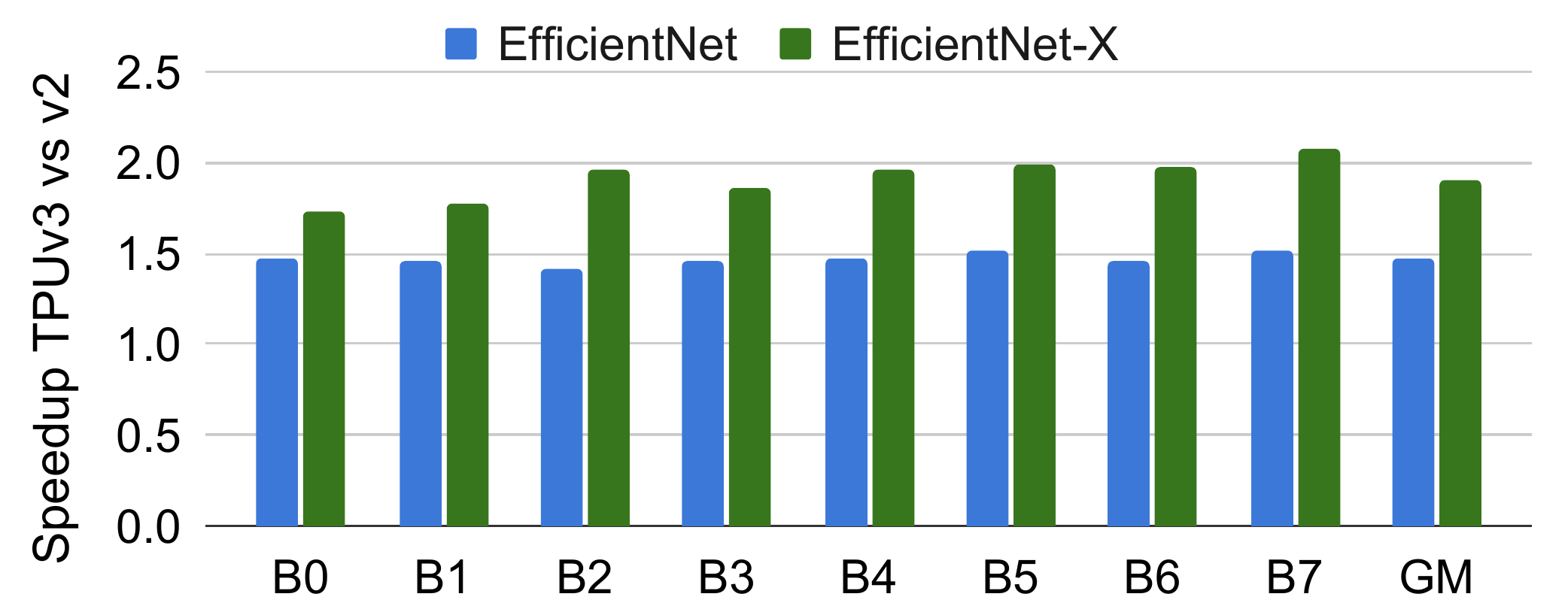}
\caption{\small Speedup of EfficientNet-X and EfficientNet when migrating from TPUv2 to TPUv3 with 2X hardware peak performance. GM is geometric mean.
}
\label{fig:speedup_tpuv2v3}
\end{figure}

Moreover, the DC accelerator-friendliness of EfficientNet-X generalizes well across accelerator generations. Specifically, as shown in Figure~\ref{fig:speedup_tpuv2v3}, TPUv3 has 2X peak performance than TPUv2. When migrating from TPUv2 to TPUv3, EfficientNet-X models achieve $\sim$1.9X average (geometric mean) speedup while EfficientNet models only achieve $\sim$1.5X speedup. In other words, EfficientNet-X materializes $\sim$30\% better speedup than EfficientNet when migrating from TPUv2 to TPUv3, demonstrating good generality.

All these results demonstrate the effectiveness of our method. Specifically, our method, including NAS with the search space optimized for DC-accelerators and LACS,  emphasizes on simultaneously optimizing total computation $W$, operational intensity $I$, and execution efficiency $E$.



\section{Related work}
\label{sec:related_work}
Neural Architecture Search (NAS) attempts to automate the design process of machine learning models with reinforcement learning~\citep{nas_cifar17,nas_imagenet18},  evolutionary search~\citep{amoebanets18}, differentiable search~\citep{diffnas18, dong2019searching}, and other methods~\citep{nao18neural,bayesian2018neural}. Recent work in NAS has also reduced search costs~\citep{enas18,pnas18,rnas18} and improved inference efficiency~\citep{mnas18, wu2019fbnet,efficientnet19, lyna2020fast, li2019partial}.
When designing fast models for inference with NAS, previous work employed multi-objective search~\citep{mnas18,elsken2018efficient,proxyless18,monas18,rnas18,mobilenetv319,cai2019once, guo2020single, lu2019nsga, dong2018ppp} to consider accuracy together with performance/efficiency. However, their methods only passively use high level signals such as model size and latency. 

Targeted ML optimizations are also used extensively to improve model accuracy and efficiency trade-offs. These targeted optimizations include automated approaches such as model pruning and quantization~\citep{prune,amc18,sscnn,sparscnn17,SparseWinograd:Liu,SparseWinograd:Li,squeezeNext18,squeezenet16,nas_cifar17} as well as manual optimizations on specific platforms especially mobile devices~\citep{mobilenetv117,mobilenetv218}. 

Initial model scaling involves taking a fixed architecture and individually increasing depth \citep{resnet16} and width \citep{wideresnet16} in separation or together~\citep{mobilenetv117, resnest2020}. Further work in compound scaling yielded model families varying in depth, width, and resolution simultaneously and systematically \citep{efficientnet19}. Scaling is also more recently used in constructing larger models in conjunction with NAS \citep{nas_imagenet18, efficientnet19}. 

Specialized datacenter accelerators have been playing a critical role in powering machine learning. 
These accelerators, including TPUs~\citep{dean2019deep,tpuv1:isca:2017} and GPUs~\cite{nv:v100,nv:a100}, provide the computing power for both training and inference at scale.\vspace{-0.01in}

\section{Conclusions}
\label{sec:conclusions}
This work presents a new method to search for CNN model families targeting datacenter accelerators for high accuracy and efficient inference. We first provide analysis to show the root cause of FLOPs-latency nonproportionality and ways to improve CNN performance on DC accelerators. Guided by the insights gained from our analysis, the new search method incorporates a NAS search space tailored for DC accelerators and a new scaling approach called latency-aware compound scaling. 
 Our new method provides the search and scaling for model families with more visibility into accelerator details, and compose model families with optimized FLOPs, operational intensity, and efficiency to achieve better accuracy and speed. Note that although we choose EfficienNet as the baseline in the paper, our method is generic and can improve efficiency for any SOTA model families on DC accelerators.  The resulted EfficientNet-X model family achieves up to 2X+ faster speed and comparable accuracy to SOTA model families on TPUv3 and GPUv100. EfficientNet-X also achieves better speedup migrating from TPUv2 to TPUv3, which demonstrates the generality of our method across different accelerator generations. These results highlight the impressive possibilities available through careful optimizations on NAS and compound scaling for increasingly demanding computer vision models on emerging DC accelerators.

\section*{Acknowledgement}
\update{We thank Aoxiang Cui, Jeff Dean, Rama Govindaraju, Samuel Kwong, Chenhao Li, David Lo, Yun Ni, Nishant Patil, David Patterson, Parthasarathy Ranganathan, Adrian Smarandoiu, Vijay Vasudevan, Shibo Wang, and Shengqi Zhu for their help on infrastructure/benchmarking and/or constructive comments on early drafts of this paper.}
\newpage
\medskip
{\small
\bibliographystyle{ieee_fullname.bst}
\bibliography{main,cv}
}

\appendix

\begin{table*}[t!]
\small
    \centering   
    \small
        \begin{tabular}{l|c|cc|ccc}           
        \small
        \multirow{2}{*}{\bf Model}	 &  \multicolumn{1}{c|}{ \bf Top-1 Accuracy (\%).$^{\star}$}	  & \multicolumn{1}{c}{\bf\#Params} & \multicolumn{1}{c|}{\bf \#FLOPs$^{\dagger}$} & \multicolumn{1}{c}{\bf I$^{\P}$} & \multirow{2}{*}{\bf E$^{\ddagger}$} & \multicolumn{1}{c}{\bf Inference Latency$^{\S}$(ms)} \\
        & (\bf  TPUv3 / GPUv100) & (\bf Million) & (\bf Billion) & (\bf Ops/Byte) &  & (\bf  TPUv3 / GPUv100)  \\
        \midrule [0.1em]                                           
        EfficientNet-B0~\cite{efficientnet19}  & 77.3  & 5.3  &   0.39 & 19.7& 52.4\% & 13.4 / 38.1\\
        \hline
        +SpaceToDepth  & 77.5  & 7.2  &   0.47 & 25.3& 55.8\% & 11.9 / 35.6\\
        \hline  
        +Fused Conv  & 77.8  & 7.6  &   0.91 & 62.5& 56.1\% & 9.5 / 30.5\\
        \hline  
        \multicolumn{1}{l|}{+Activation}  & \multicolumn{1}{c|}{77.4 / 77.3}  & \multicolumn{1}{c}{7.6}  &   \multicolumn{1}{c|}{0.91} & \multicolumn{1}{c}{63.8} & \multicolumn{1}{c}{57.3\%} & \multicolumn{1}{c}{8.7 / 22.5}\\
        (EfficientNet-X-B0) & ~ & ~& & \\
    \end{tabular}  
     \caption{                                                                      
         Contribution breakdowns of each enhanced model architectures to inference latency and accuracy on imagenet of the searched based model EfficientNet-X-B0 on TPUv3 and GPUv100. TPUv3 and GPUv100 results are separated by "/" when they differ, shown as "TPUv3 results / GPUv100 results". $^{\star}$Only with the different activation function selections, accuracies differ on TPUs and GPUs. $^{\dagger}$Following common practices, \#FLOPs refer to \#multiply-and-add operations. $^{\P}$I is the operational intensity measured on TPUv3. $^{\ddagger}$E is the execution efficiency measured on TPUv3, w.r.t to roofline instead of peak hardware FLOPs/sec as shown in Equation~\ref{eqn:optimization}. Only in the compute-bound region as shown in Figure~\ref{fig:roofline_overall}, the roofline and peak hardware FLOPs/sec are the same. $^{\S}$The inference latency are measured for inferencing 128 images on TPUv3 and GPUv100, with mini batch size of 128. Models run in FP16 mode on GPUv100.   
       } 
    \label{tab:base_model_breakdown}     
\end{table*}

\section{Ablation study on the DC accelerator optimized search space and the searched EfficientNet-X-B0 base model}
\label{sec:search-ablation}

As summarized in Section~\ref{subsec:nas_search} and Section~\ref{sec:experiments}, all enhancements in the DC-accelerator-optimized search space (Section~\ref{sec:search_space_nas}) contribute to improving accuracy-latency trade-offs in the searched base model --- EfficientNet-X-B0. Table~\ref{tab:base_model_breakdown} shows the detailed ablation study on how these new model architecture components, including space-to-depth, fused convolution structures, and block-wise searchable activation functions, improve accuracy-latency-Pareto results over the baseline EfficientNet-B0. The Space-to-depth and fused convolution structures improve both the accuracy and speed on TPUv3 and GPUv100. The trends on the total FLOPs further confirms our analysis on new search space about activation functions as described in Section~\ref{sec:search_space_nas} and Section~\ref{subsec:nas_search}. Concretely, although activation functions have negligible impact on total model FLOPs on TPUs and GPUs, they have big impact on performance. On GPUv100, NAS selects ReLU activation for all layers/blocks for EfficientNet-X-B0 because of the performance degradation caused by non-fused swish. On TPU, NAS selects ReLU for blocks with depthwise convolutions and swish for blocks with vanilla convolutions to avoid overloading the vector units in TPUv3 as described in Section~\ref{subsec:nas_search}. As a result, the new activation function strategy improves speed but causes accuracy drop on both GPUv100 and TPUv3. However, thanks to the accuracy improvements from space-to-depth and fused convolutions, the final accuracy is comparable to the baseline EfficientNet-B0 on both TPUv3 and GPUv100 as shown in Table~\ref{tab:base_model_breakdown}. The hybrid ReLU and swish activation functions on TPUv3 leads to the higher accuracy than the ReLU-only activation functions on GPUv100. Note that in Table~\ref{tab:imagenet}, we report the lower accuracy from TPUv3 and GPUv100 as the final score. 

On TPUv3, all new enhanced search space components contribute almost equally in inference speed, with the new activation function strategy offsetting some of the accuracy gains. On GPUv100, the new activation function strategy causes a more significant inference speedup than other new model architecture enhancements, but with a bigger accuracy drop than on TPUv3. This demonstrates the impact of the software stack. We believe a fused swish implementation for GPU software will make GPUv100 behave similar to TPUv3. \looseness=-1 


\section{Ablation study on latency-aware compound scaling and the EfficientNet-X family}
\label{sec:lacs-ablation}
\begin{table}[t!]
    \centering
    \resizebox{1.0\columnwidth}{!}{
    \begin{tabular}{c|c|c}
        LACS search level & $\dagger$Coefficients $\alpha, \beta, \gamma$ & X-B7 Dimensions \\
        \midrule [0.1em] \\
        LACS at X-B1 level & (1.27, 1.16, 1.08) & (Depth: 75, Res: 368)\\
        LACS at X-B7 level & (1.28, 1.17, 1.07) & (Depth: 79, Res: 350)\\
    \end{tabular}
    }
        \caption{Scaling coefficients $\alpha, \beta, \gamma$ and model dimensions yielded by LACS at low (X-B1, \ie, EfficientNet-X-B1) level and directly at (X-B7, \ie, EfficientNet-X-B7) level on GPUv100. $\alpha, \beta$, and $\gamma$ are the base exponential terms to be used together with $\phi$ as described in Equation~\ref{eq:optobj}. Depth means total number of layers in the network. Res means the input resolution. Both scaling coefficients and model dimensions (depth, input resolution) produced by the methods are quite similar. }
    \label{tab:lacs-ablation}
\end{table}

\begin{table}[t!]
    \centering
    \resizebox{1.0\columnwidth}{!}{ 
    \begin{tabular}{c|c|c|c}
         \multicolumn{1}{c|}{EfficientNet-X } &  \multicolumn{1}{c|}{Single-obj } &  \multicolumn{1}{c|}{LACS } & \multicolumn{1}{c}{LACS } \\
         Model &  scaling & on GPU & on TPU\\
        \midrule [0.1em] \\
        X-B0 & (16, 224) & (16, 224) & (16, 224)\\
        X-B1 & (17, 240) & (17, 229) & (17, 229)\\
        X-B2 & (19, 260) & (20, 241) & (20, 243)\\
        X-B3 & (22, 300) & (25, 258) & (26, 263)\\
        X-B4 & (28, 380) & (36, 289) & (38, 298)\\
        X-B5 & (35, 456) & (49, 317) & (52, 331)\\
        X-B6 & (41, 528) & (62, 343) & (68, 361)\\
        X-B7 & (49, 600) & (79, 368) & (87, 391)\\
    \end{tabular}
    }
    \caption{Comparison on depth (\ie, layer count of the network) and input image resolution of EfficientNet-X model family with different compound scaling factors designed by LACS and single-objective compound scaling. Each result contains a pair of "(depth, input image resolution)". since single-objective compound scaling only uses accuracy as the sole objective, it does not produce different scaling factors for TPUv3 and GPUv100. The base model EfficientNet-X-B0 is also included, which is the same for all cases.}
    \label{tab:lacs-family-dimensions}
\end{table}
As summarized in Section~\ref{subsec:scaling_lacs} and Section~\ref{sec:experiments}, LACS achieves a better set of scaling factors than single-objective compound scaling that originally proposed in the EfficientNet~\cite{efficientnet19} work. Clearly, searching for scaling coefficients at a lower target latency level (\eg, EfficientNet-X-B1) and using them to create higher latency models (\eg, Efficientnet-X-B7) is much more cost-effective than directly searching for coefficients at the higher latency model level (\eg, Efficientnet-X-B7). However, searching first at low latency level models and scaling to high latency level models has the potential to deviate from the empirical optimum from direct searching at high latency level models, due to non-linear increases of accuracy and latency with larger depth, width, and resolution. In this ablation study, we first verify the efficacy of LACS in maintaining good scaling from small to large models, without deviation from the empirical optimum. We then provide more comparisons on results from LACS and single-objective compound scaling. 

To verify the efficacy of LACS, we target the B7 model level of the EfficientNet-X family on GPU and compare the scaling factors yielded by LACS at X-B1 level and then applied at X-B7 level against direct accuracy-latency-Pareto-search at the X-B7 level to find the empirical optimum coefficients. As shown in Table~\ref{tab:lacs-ablation}, both the scaling coefficients and the resulting network dimensions are quite similar. Particularly, the network dimensions are within 6\% of each other. This verifies that LACS can effectively scale up all the way to the high end models to form a model family, with negligible deviations from empirical optima. 

With the verified efficacy of LACS, we present detailed comparisons on model dimensions of EfficientNet-X on TPUv3 and GPUv100 with the scaling factors obtained by LACS and by the original single-objective compound scaling as used in EfficientNet~\cite{efficientnet19}. We first run single-objective compound scaling that uses accuracy as the sole objective as proposed in~\cite{efficientnet19}. Even with the new EfficientNet-X-B0 as the base model, the single-objective compound scaling method finds the same compound scaling factors as with EfficientNet. On the other hand, LACS finds different compound scaling factors on TPUv3 and GPUv100. Table~\ref{tab:scaling-coefficients} shows these different scaling factors obtained from LACS and single-objective compound scaling. Note that since single-objective compound scaling only uses accuracy as the sole objective, unlike LACS, it does not generate different scaling factors for TPUv3 and GPUv100. Table~\ref{tab:lacs-family-dimensions} shows the detailed model dimensions generated by these different scaling factors. While LACS creates different families for TPUv3 and GPUv100, the most notable difference is that both LACS versions prefer deeper and slimmer models as compared to original single-objective compound scaling, with the LACS results on GPU and TPU being 60\% $\sim$ 70\% deeper with $\sim$40\% smaller input resolutions. The changes in scaling and the resulted model architectures are caused by the use of the accuracy-latency multi-objective that provides more visibility into the hardware architecture details. As a result, EfficientNet-X has much faster inference speed, with comparable accuracy to EfficientNet as shown in Table~\ref{tab:imagenet} and Figure~\ref{fig:speedup_nas_lacs}.


\end{document}